\documentclass{article}
\usepackage{ijcai18}

\usepackage{times}
\usepackage{xcolor}
\usepackage{soul}
\usepackage[utf8]{inputenc}
\usepackage[small]{caption}
\usepackage{amsmath}

\usepackage{listings}
\usepackage{booktabs}
\usepackage{graphicx}

\title{Diverse Beam Search for Increased Novelty in Abstractive Summarization}

\author{
Cibils André$^\dagger$, 
Musat Claudiu$^*$,
Hossmann Andreea$^*$,
Baeriswyl Michael$^*$,
\\ 
$^\dagger$ École Polytechnique Fédérale de Lausanne (EPFL) \\
\textit{Email: } \{\textit{firstName.lastName}\}@epfl.ch \\
$^*$ Artificial Intelligence Group - Swisscom AG\\
\textit{Email: } \{\textit{firstName.lastName}\}@swisscom.com 
}

\begin{document}

\maketitle

\begin{abstract}

Text summarization condenses a text to a shorter version while retaining the important informations. Abstractive summarization is a recent development that generates new phrases, rather than simply copying or rephrasing sentences within the original text.

Recently neural sequence-to-sequence models have achieved good results in the field of abstractive summarization, which opens new possibilities and applications for industrial purposes. However, most practitioners observe that these models still use large parts of the original text in the output summaries, making them often similar to extractive frameworks.

To address this drawback, we first introduce a new metric to measure how much of a summary is extracted from the input text. Secondly, we present a novel method, that relies on a diversity factor in computing the neural network loss,  to improve the diversity of the summaries generated by any neural abstractive model implementing beam search.  
Finally, we show that this method not only makes the system less extractive, but also improves the overall rouge score of state-of-the-art methods by at least 2 points.
\end{abstract}

\section{Introduction}
Summarization is a process of generating a condensed version of a text that contains the key information from the original. For automatic summarization, two approaches can be used: extractive and abstractive, sometimes also called generative. 

Extractive summarization is focused on finding the most relevant text spans or phrases in the original document and to copying them to construct a summary from it. The final text consists exclusively of passages from the input. This process can thus be viewed as a ranking mechanism~\cite{Jin2010}.

A second summary generation paradigm is to generate novel text to produce the summary. This generative process sometimes uses words coming from a vocabulary unseen in the source document, as a human usually does. Abstractive summarization is considered more difficult because it requires high-level competences such as generalization or reformulation, compared to extractive summarization which ensures baseline levels of grammaticality and accuracy. 

Until the advent of powerful deep architectures, such as Generative Adversarial Networks for text generation~\cite{Rajeswar2017,Zhang2017} and pointer networks~\cite{Schmidhuber2015}, the generative summarization was largely considered infeasible. These novel methods are now opening the way to a new generation of summarization systems that carry the promise of near-human text generation quality. 
Practitioners however quickly discarded the earlier generative models due to their inability to solve problems that make the result look robotic and incomprehensible - chiefly among them repetition.

A recent model, the Pointer Generator Network~\cite{See2017} or PGNet, achieves state of the art in term of automatic summarization on a widely used dataset - the CNN/Dailymail~\cite{Hermann2015,Nallapati2016,See2017}. This model is an abstractive sequence-to-sequence neural model that uses both extractive and generative summarization techniques. To overcome problems that plagued previous sequence-to-sequence neural models, like repetition and the inability to handle out of vocabulary (OOV) words, the PGNet features several improvements. These include attention and a hybrid mechanism that enables the model to copy words from the original text - a pointer network~\cite{Schmidhuber2015}. 

The inclusion of a pointer generation step has a clear positive impact on the performance of the system. This progress is evident both quantitatively (e.g. ROUGE scores) and qualitatively~\cite{See2017}. There is a caveat though. 
Due to its extractive pointer generator component, the model has the option of copying large swaths of the original text. By overusing this option, the method becomes self-defeating and leads to an overly extractive behavior in an abstractive method.
We argue that this drawback is inherently common across summarizers that use pointers to elements in the original text.

In this work, we propose a way to measure the extractiveness of abstractive summarization methods. We show that common methods used in plagiarism detection cannot identify cases where the summarizer simply pieces together several large text spans from the original text. We thus devise a novel method to penalize this behaviour and use it as a complement to traditional evaluations like ROUGE-N and ROUGE-L scores. The goal of this orthogonal evaluation is to discourage abstractive models from going for the low hanging fruit - copying.

A second contribution is a method to reduce the extractiveness of abstractive summaries, based on a diverse beam search (DBS)~\cite{Vijayakumar2016}. 

A previous attempt \cite{Li2016} to improve abstractive summaries with DBS had lacklustre results. While each diverse summary is by itself not superior to the one obtained by a baseline model, we propose a method of combining the novelty within various DBS-based summaries.
We show that the DBS-based results are both qualitatively and quantitatively better the the previous state of the art. A major result is that we can thus decrease the extractive nature of the summarizer while at the same time increasing the ROUGE-1, ROUGE-2 and ROUGE-L scores on the CNN/Daily Mail dataset~\cite{Hermann2015}.
The method is compatible with any abstractive summarizer that uses beam search, which includes the leading methods at the time of the writing~\cite{Nallapati2016,See2017,Paulus2017,Chopra2016,Hasselqvist2017}.

\section{Related Work}

Until recently, text summarization has been in a vast majority extractive~\cite{Wong2008,Chuang2000}. We are however on the cusp of a major paradigm shift towards abstractive summarization, largely due to the good results obtained by recent models.

The first work that succeed in doing abstractive summarization using a sequence-to-sequence model was \cite{Nallapati2016}, who introduced a major dataset CNN/Daily Mail, that we also use in the current work. They addresses multiple issues such as capturing the hierarchy of sentence-to-word structure, and emitting words that are rare at training time. Independently, \cite{Chopra2016} created an attentive recurrent model that yielded good results on the DUC-2004 shared task. More recently, \cite{Paulus2017} created a reinforcement model for abstractive summarization that became a new state-of-the-art on the CNN/Daily Mail dataset.

This surge in interest in abstractive summarization from the research community thus lead to increasingly promising results. In practice however, there are remaining issues that preclude their use in industrial settings: inaccurately reproducing factual details, an inability to deal with out-of-vocabulary (OOV) words and repetition. The PGNet~\cite{See2017} was designed as an architecture to tackle  all of these issues simultaneously by pointing to and retrieving elements from the original text.

While it successfully alleviates the impact of the aforementioned ills, the PGNet tends to abuse its copying mechanism and the generated summaries are largely extractive.

To address this novel problem, we first need to quantify it. We define the extractiveness of an abstractive summarizer as how much of the extracted summary is copied from the original text. The extractiveness is thus akin to plagiarism detection, which measures how much a text is extracted from another.
Plagiarism detection is a well studied problem and two frequently used measures are: n-grams frequencies and longest common sequence (LCS)~\cite{Zhang2012,Anzelmi2011}. 
These techniques are not adequate measures of the extractiveness of a summary, as we can expect from a summary to share a lot of words with the document.

The solution we present to reduce the extractiveness is based on diverse beam search (DBS)~\cite{Vijayakumar2016}. DBS was shown to be effective for creating diverse image captions, machine translation and visual question generation. It is a variation of the classic beam search designed for neural sequence models which addresses the lack of diversity of the original algorithm~\cite{Gimpel2013}.

DBS has been used in multiple topics, such as dialogue response generation~\cite{Asghar2016}, machine translation\cite{Li2016}, but also abstractive summarization \cite{Li2016}. However, DBS on its own contributes only marginally (+0.25 $F_1$-score) to the performance of abstractive summarization. This is why we combine it with a candidate selection algorithm used in multiple fields, Maximal Marginal Relevance (MMR)~\cite{Guo2010,Carbonell1998}. %MMR
MMR is an algorithm that balances relevance and diversity in multiple set-based information retrieval tasks.

In order to use MMR, one needs to compute the similarity between two sentences. Many options exist, but recently good results have been obtained using sentence embeddings, for instance ones based on n-gram features (Sent2Vec)~\cite{Pagliardini2017}. Sent2vec produces specific word and n-gram embedding vectors, which are additively combined into a sentence embedding. Similarly to word embeddings~\cite{Mikolov2013}, Sent2Vec allow us to represent semantic relatedness between phrases.

\section{Summary Generation}
The first part of this section is devoted to briefly describing the baseline model, PGNet~\cite{See2017}. 
In the second half we describe one of our main contributions - how to enhance PGNet in order to generate less extractive summaries.

\subsection{Baseline Model}
The core idea behind current methods for text summarization consists in leveraging a corpus containing both the source documents and their summaries.
PGNet, in particular, learns how to map an input sequence of words (the source document) to another sequence of words (the summary of the document) by implementing the well-known sequence-to-sequence neural network architecture, extended with an attention mechanism~\cite{Bahdanau2014}.
Additionally, PGNet addresses some of the shortcomings of previous models, such as the inability to handle out-of-vocabulary words and repetition~\cite{Sankaran2016,Tu2016}.

The main novelty of PGNet is the Pointer Network~\cite{Schmidhuber2015} that aggregates the context vector produced by the attention mechanism and the decoder state, making the model able to ``copy'' words from the original document, and to combine them with the fragments of the output summary generated in an abstractive fashion.
It is thus possible for the model to make use of words that are not contained in its vocabulary (i.e., they were never seen during the training phase) but that appear in the input text.
Concretely, when computing the next word composing the summary, a probability $p_{t} \in [0,1]$ is generated based on the context vector, the decoder state and the decoder input. 

Such probability is then used as a soft switch to choose between generating a word from the known vocabulary, or extracting a word from the original text.

Finally, as it is common practice in many tasks such as Machine Translation and Text Summarization, PGNet uses the Beam Search algorithm to generate its summaries. 
When generating each word composing the summary, instead of greedily taking the word with the highest score, the algorithm selects the top-$B$ best scoring candidates, thus exploring multiple sequences in parallel.

Nevertheless, what often happens in practice is that the output generated by such heuristic tend to stem from a single highly valued beam, resulting on minor perturbations of a single sequence.

\begin{figure}[t!]
\centering
\includegraphics[width=.49\textwidth]{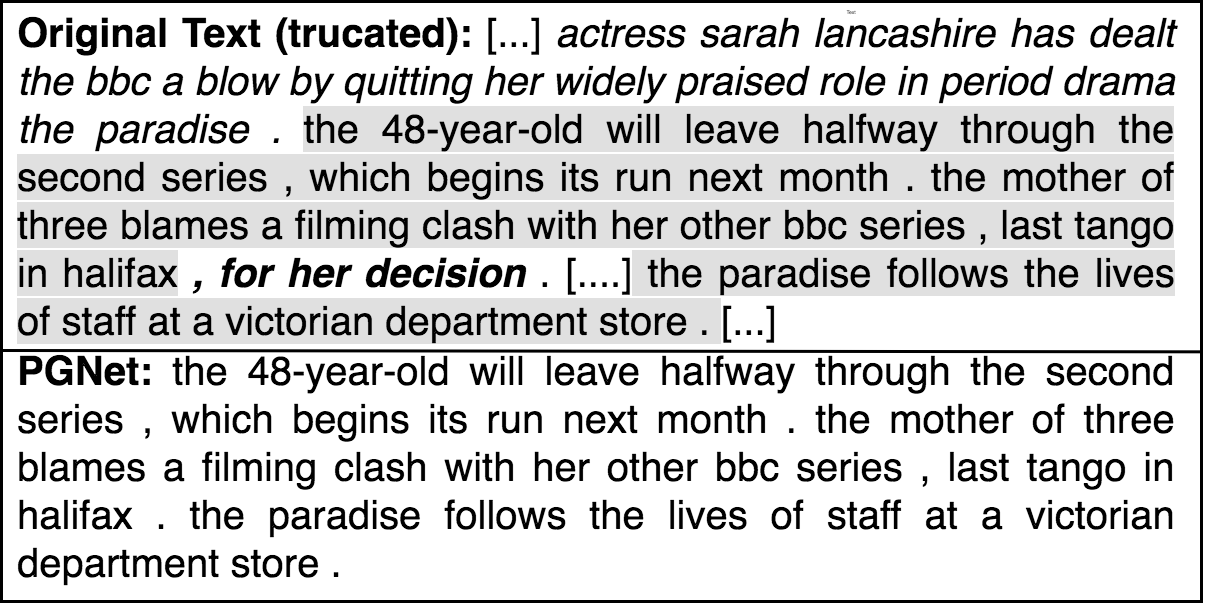}
\caption{A summary generated by the baseline model, the PGNet. Although it is overall extractive, we see that some parts have been cut of.}
\label{subfig:pgnet_extractive_example}
\end{figure}

\paragraph{PGNet in Practice}
In practice, PGNet acts very much as an extractive model: as ~\cite{See2017} point out in their work, 35\% of the time the model copies whole article sentences, meaning the model behave as fully extractive. In the others 65\%, the model encompasses a range of abstractive techniques, such as truncating sentences to grammatically-correct shorter versions, which still make the model feel extractive~\cite{See2017}.
This can also be seen in Figure~\ref{subfig:pgnet_extractive_example}, in which we highlighted the portion of the input text that is reused in the summary.

\subsection{Extractiveness}

\newcommand{\plagiarism}{\mathop{\mathrm{plagiarism\_score}}}
\newcommand{\extraction}{\mathop{\mathrm{extraction\_score}}}

These observations motivated the current work. To tackle extractiveness, a first step is to measure it.
Some methods already exist for simple plagiarism detection, based on n-gram frequencies analysis or longest common sequence~\cite{Zhang2012,Anzelmi2011}. From LCS, we can easily define a $\plagiarism$ by normalizing the length of the LCS between a given summary and a document by the length of the summary.
However, this metric has a major flaw: it completely discards the rest of the summary, and thus make no difference between a summary constructed from two sequences from the original document and a summary that is half composed of new words outside its LCS with the article. 

To address this issue, we define a metric which takes more parameters than just LCS into account and more generally penalizes any large spans of text copied from the original text.
Merging sentences or splitting them should not be penalized as heavily as just copying them wholly. Moreover, we want the metric to be consistent, deterministic and to output a normalized score. Finally, novel word combinations should not be penalized at all.

We thus define an $\extraction$ as:
\begin{equation}
\extraction(S) = \sum_{s \in P(ACS_S)} s \times (e^{s-1} - (1-s)/e)
\end{equation}

Where $ACS_S$ is the set of all long non overlapping common sequences between the summary $S$ and the document, and $P(ACS_S)$ is the set of the proportion of these common sequences, i.e. each element of this set is the length of a common sequence divided by the length of the summmary.

We first start by finding the long common sequences between the summary and the original article. For this, we use a similar algorithm to LCS which will return all long non overlapping common sequences. 

This $\extraction$ ensures that the sum of these scores is between 0 and 1. Moreover, having two distinct common sequences of text of proportion $p_1$ and $p_2$ will be less penalized than having one common sequence of text of proportion $p_1 + p_2$.

A summary which consists of only one long common sequence with the document will have a score of 1, while a summary with only new words not encountered in the article will have a score of 0.

On the reference summaries, the average $\plagiarism$ and $\extraction$ over the test set are respectively $0.16$ and $0.05$. This means that the reference summaries are mainly paraphrasing and not copying text from the article, as expected. It also outlines the value of the extractiveness measure, as the goal of the automated systems is to be close to zero, as is the case for the human standard. The $\extraction$ allows us to compare different system and rank them with respect to their generative capacity.

\section{Improved Decoding Mechanism}

Our solution to the extractivness problem, is complementary to the baseline architecture, and could be used on any abstractive model that employs a beam search.

\subsection{Diverse Beam Search}
Beam search is an iterative algorithm widely used to decode RNNs~\cite{Nallapati2016,See2017,Paulus2017,Chopra2016,Hasselqvist2017} that approximates the optimal solutions. At each time step, the model computes $Y_{t}=\{y_{1,t}, ..., y_{B,t}\}$ the set of B solutions held at the start of the $t+1$ time-step:
\newcommand{\argmax}{\mathop{\mathrm{arg\,max}}}
\begin{equation}
Y_{t} = \underset{y_{1,t}, ..., y_{B, t} \in{V_t}}{\argmax} \quad \sum_{b \in{[B]}} \Theta(y_{b,t}) \quad s.t. \quad y_{i,t} \ne y_{j,t}
\end{equation}
Where $B$ is the beam width, $\Theta(y_{i,t})$ the log probability of a partial solution, $V$ the vocabulary and $V_t =  Y_t \times V$ the set of all possible token extensions of the beams $Y_t$ from which BS will pick $B$ elements from.\\

Diverse beam search on another hand decodes diverse lists by dividing equally the beam size between groups, allocating to each of them the same beam budget, i.e. the number of nodes expanded at each time-step for a given group, and enforcing diversity between groups of beams:
\begin{equation}
Y^g_{t} = \underset{y^g_{1,t}, ..., y^g_{B', t} \in{V^g_t}}{\argmax} \quad \sum_{b \in{[B']}} \Theta(y^g_{b,t}) + \lambda \Delta(Y^1_t, ..., Y^{g-1}_t)[y^g_{b,t}]
\end{equation}
$$
s.t. \quad y_{i,t} \ne y_{j,t} \quad \lambda \geq 0
$$

where $\Delta(Y^1_t, ..., Y^{g-1}_t)[y^g_{i,t}]$ is the diversity term measuring the dissimilarity of group $g$ against prior groups if token $y$ is chosen to extend any of the beams in the group $g$, and $\lambda$ the diversity strength.

By applying a variable diversity strength and a diversity measure, we observed that forcing the model to generate multiple and diverse summaries logically pushed the model to improvise more and use less of its extractive capabilities.
Many different diversity terms can be used and we present two examples in the following section.

\subsection{Merging Diverse Summaries}
The summaries based on DBS are not intrinsically better than the summaries generated by the classic beam search, but they contain more novelty.
From this, we inferred that generating multiple summaries and then picking and merging the best sentences of these summaries could lead to not only a less extractive model, but also one that is better at capturing the relevant aspects. We thus split all the diverse summaries into sentences, which become candidates to form part of a good and diverse summary.

To pick the best sentences, we first rank them. We use a framework which gave good results in keyphrase extraction with sentence embeddings~\cite{Bennani-Smires2018}.
The method is based on Sent2Vec~\cite{Pagliardini2017}, which allows the embedding of arbitrary-length sequences of words. Similarly to word embeddings, it can be used to represent semantic relatedness between phrases by using standard similarity measures, like cosine or Euclidean.

We generate a document embedding using Sent2Vec. This is done by simply concatenating all the sentences from the document and subsequently treating the document as a single phrase. The document embedding is then used to rank the usefulness of candidate sentences. 

The intuition is that the most useful candidate phrases are at the same time close to the document and far away from each other.
We thus use Maximal Marginal Relevance (MMR) to pick the candidates. MMR is used in information retrieval~\cite{Guo2010,Carbonell1998}, and balances relevance - in our case similarity to the original document to summarize - and diversity.
We compute the cosine similarity between the candidates' embedding and the document embedding to obtain a score which measures how much relevant information the candidate phrase contains. 

More precisely, we pick $N$ candidates iteratively using MMR:
\newcommand{\cossima}{\mathop{\mathrm{sim_1}}}
\newcommand{\cossimb}{\mathop{\mathrm{sim_2}}}
\begin{equation}
MMR:= \underset{C_i \in C \setminus K}{\argmax} [\beta \cossima(D, C_i) - (1-\beta) 
\underset{C_j \in K}{\max}
\cossimb(C_j, C_i)]
\end{equation}

Where $C$ is the set of candidates, i.e. the sentences, $K$ is the set of already picked candidates, $C_i$ and $D$ are the embeddings of candidate $i$ and of the document, respectively, $\cossima$ and $cossimb$ are a similarity measure, here a normalized cosine similarity as described below, and $\beta$ is a trade-off parameter between diversity and classic ranking. The choice of $N$ and the diversity factor $\beta$ is detailed in the following section.

We define $\cossima$ as below:

\begin{equation}
ncos_{s}(D, C_i):= \dfrac{cos_{sim}(D, C_i)}{\underset{C_k \in C}{\max}cos_{sim}(D, C_k)}
\end{equation}

\begin{equation}
\cossima(D, C_i):=0.5 + \dfrac{ncos_{s}(D, C_i) - \overline{ncos_{s}(D, C)}}{\sigma(ncos_{s}(D, C))}
\end{equation}

Where $D$ is the document embedding, $\overline{ncos_{sim}(D, C)}$ and $\sigma(ncos_{sim}(D, C))$ represent the average similarity and the standard deviation between $D$ and the set of candidates $C$.

We apply the same kind of transformation for the similarity among the candidate phrases themselves, as shown by the following equations for $\cossimb$:
\begin{equation}
ncos_{s}(C_j, C_i):= \dfrac{cos_{sim}(C_j, C_i)}{\underset{C_k \in C \setminus \{C_i\}}{\max}cos_{sim}(C_i, C_k)}
\end{equation}

\begin{equation}
\cossimb(C_j, C_i):=0.5 + \dfrac{ncos_{s}(C_j, C_i) - \overline{ncos_{s}(C_i,C \setminus \{C_i\})}}{\sigma(ncos_{s}(C_i, C \setminus \{C_i\}))}
\end{equation}

By selecting the best sentences, we can construct the final summary which contains the best elements of the generated diverse summaries.

\section{Experiments}
We evaluate the baseline PGNet and our diverse generative model using the same pyrouge package\footnote{https://pypi.python.pyrouge.0.1.0} as ~\cite{Nallapati2016,See2017}. 
For all experiments, we use a trained PGNet as described in the original paper~\cite{See2017} with coverage using 256-hidden states for the encoder and the decoder LSTMs and 128-dimensional words embeddings. The words embedding are not pretrained but are learned during training.

We train using Adagrad~\cite{Duchi2011} with learning rate 0.15 and an initial accumulator value of 0.1. Gradient clipping is used with a maximum gradient norm of 2, and there is no regularization. 
During training, we also truncate the article to 400 tokens, and for generating new summaries, the limit length of the summary is at first set to 120, similarly to the original model.
\begin{figure*}[h!]
\centering
\includegraphics[width=0.95\textwidth]{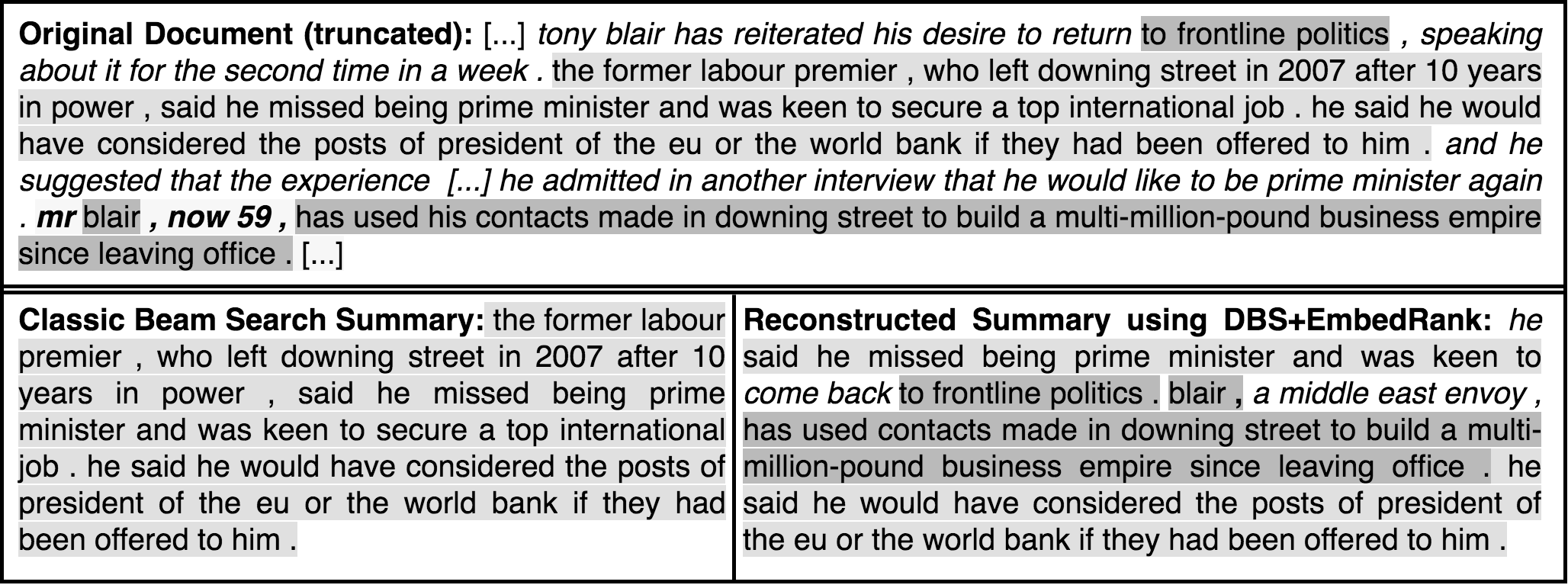}
\caption{Two summaries generated by PGNet with Beam Search and by PGNet with DBS+EmbedRank. Although the baseline model behave as a fully extractive one, our post processing module output a more diverse and abstractive summary.}
\label{subfig:other}
\end{figure*}

\begin{table*}
\centering
\begin{tabular*}{\textwidth}{|l @{\extracolsep{\fill}} |c|c|}
\hline
 & $\plagiarism$ & $\extraction$ \\ \hline
Pointer-Generator Network with Beam Search decoder (baseline) & 0.566 & 0.401 \\ \hline
Pointer-Generator Network with DBS+EmbedRank decoder, $N$ = 3, $\beta$ = 0.35 \qquad \qquad & 0.440 & 0.282 \\ \hline
Pointer-Generator Network with DBS+EmbedRank decoder, $N$ = 4, $\beta$ = 0.20 \qquad \qquad & \textbf{0.368} & \textbf{0.211} \\ \hline
Reference summaries & 0.158 & 0.049 \\ \hline
\end{tabular*}
\caption{$\plagiarism$ and $\extraction$ on the test set. Except for $N$ and $\beta$, the hyper parameters stay the same for the Pointer-Generator Network. We also display the score from the reference summaries for comparison.}
\label{table:results_extraction}
\end{table*}

\subsection{Hyper-Parameter Description and Selection}

\subsubsection{Best Set of Hyper-Parameters}
Testing exhaustively each of these hyper-parameter is expensive, thus we performed the experiments on a small portions on the dataset (3500 randomly picked examples on the training set) on multiple sets of hyper parameters. Using ROUGE-1 and ROUGE-L, we then ranked the sets of hyper parameters and picked the best set. This set is then tested on the original test dataset to confirm that no bias is hidden in this small dataset.

\subsubsection{Hyper-Parameter Description}
For the Diverse Beam Search, we enforce a maximum and minimum number of tokens per summary (150 and 35). Experiments show that these are not the most important factors, as the second and third best set of parameters were just variations of that. It is important to note that the baseline PGNet uses a maximum number of tokens per summaries of 120 during testing.

The beam width $B$ is the parameter used to define the number of nodes expanded at each time-step during the search. A high value for $B$ means that the search-space is larger, which is more computationally expensive. We set $B$ to 24, which is significantly higher compared to the PGNet, where it was set to 4. A high beam size means more resources to explore each of the groups, which leads in turn to better candidates.

There is a trade-off between $B$ and the number of groups $G$. Each group requires resources and if $G$ is set to 1, it reduces DBS to BS. On the other hand setting it to the beam size $B$ allows for the maximum exploration of the search-space, in opposition to having a high budget for each group. We set the group number $G$ to 6, with a beam width of 4 per group. This allow the model to perform evenly to the baseline on the first group. This also generates 6 different summaries.

\begin{table*}
\begin{tabular*}{\textwidth}{|l @{\extracolsep{\fill}} |c|c|c|}

\hline
 & ROUGE-1 & ROUGE-2 & ROUGE-L \\ \hline
Pointer-Generator Network with a Beam Search decoder (baseline model) & 38.02 & 16.26  & 34.77 \\ \hline
Pointer-Generator Network with DBS+EmbedRank decoder, $N$ = 3, $\beta$ = 0.35 \qquad \qquad & \textbf{40.19} & 17.09 & \textbf{36.63} \\ \hline
Pointer-Generator Network with DBS+EmbedRank decoder, $N$ = 4, $\beta$ = 0.20 \qquad \qquad & 39.52 & \textbf{17.16} & 36.21 \\ \hline
\end{tabular*}
\caption{ROUGE $F_1$ scores on the test set. All of our ROUGE scores have a 95\% confidence interval of at most $\pm$ 0.25 as reported by the official ROUGE script. Except for $N$ and $\beta$, the hyper parameters stay the same.}
\label{table:results_rouge1_L_Ext}
\end{table*}

The diversity strength $\lambda$ is a scalar between 0 and 1 which penalizes summaries that look alike. More precisely, this specifies the trade-off between the joint probability and the diversity terms. A high value $\lambda$ produces more diverse summary, but excessively high values of the diversity strength can result in grammatically incorrect outputs as it can overpower model probability. 

Finally for beam search, $\Delta$ is defined as a function that outputs a vector of similarity scores for potential beam completions. Many options exist for a diversity function $\Delta$.
\begin{itemize}
\item Hamming Diversity: The current group is penalized for producing the same words at the same time. More precisely, we compute the hamming distance between two strings and normalized it over the length of the string and over the group.
\item N-Grams Diversity: The current group is penalized for producing the same n-grams as previous group, regardless of alignment in time.
\end{itemize}

We use the Hamming diversity with a diversity strength $\lambda$ of 0.3, meaning that we penalizes the selection of tokens used in previous groups proportional to the number of times it was selected before. It ensures that different words are used at different times, which forces the model to rephrases it sentences and add new words.

Then, we define the hyper-parameters for EmbedRank, which are the number of candidates to select $N$ and the diversity factor $\beta$ for the MMR. More precisely, we define $N$ as the number of iteration that we run on the MMR algorithm and thus the number of sentences in our final summary. The factor $\beta$ is a trade-off between a standard, relevance ranked list and a maximal diversity ranking of the candidates.

Finally, the parameters for EmbedRank are set to $N=3$ and $\beta=0.35$ for ROUGE-1 and ROUGE-L optimization, meaning that each summary will be three sentences long.

\subsection{DBS Improvements}

\subsubsection{Extractiveness}
We first evaluate the impact of the diverse generative summarization through the prism of the extractiveness of the resulting models. In Figure \ref{subfig:other} we portray the original PGNet and diverse summaries side by side. It is immediately obvious to a human that the latter is less extractive and contains shorter excerpts from the original text. To quantify this finding, we compute the previously discussed $\plagiarism$ and $\extraction$ for each method. The results are shown in Table \ref{table:results_extraction}.

We portray two parameter combinations for the diverse summary generator:  ($N$ = 3, $\beta$ = 0.35) and ($N$ = 4, $\beta$ = 0.20). 
Firstly, we see that both models perform considerably better with regards of both metrics. A $\plagiarism$ of $0.440$ means than in average, our summaries are less than half extracted from the text. Moreover, having a value $0.282$ instead of $0.401$ value on the $\extraction$ means that the model is in its core less extractive, even if it keeps some common sequence with the original text.

Secondly, we notice that we can reduce the extractiveness, respectively getting $0.368$ and $0.211$ on $\plagiarism$ and $\extraction$ with our model by having slightly \textbf{longer} summaries. This is a counterintuitive result that is in fact easily explainable. When picking a lower number of sentences, the impact of the diversity is in fact reduced, whereas for higher selected candidate counts the relevant information is spread across multiple sentences.

\subsubsection{ROUGE Scores}
We compare the ROUGE results for the diverse summaries with two different parameter combinations in Table \ref{table:results_rouge1_L_Ext}.

All of the summaries are evaluated using the standard ROUGE metric, more precisely the $F_1$ scores for ROUGE-1, ROUGE-2 and ROUGE-L, which measures respectively the word-overlap, bigram-overlap and the longest common sequence between the reference summary and the generated summary.

Both the summaries containing $N$ = 3 and $N$ = 4 sentences perform significantly above the baseline, with F1 score gains exceeding 2 points. 
In addition we observe that there is a slight tradeoff with the extractiveness measure. A much lower extractiveness can be obtained for a  marginal cost in term of ROUGE score.

A notable result is that the ROUGE-1, ROUGE-2 and ROUGE-L $F_1$ score improvements are obtained simultaneously with the extractiveness decrease, making the diverse symmary generation a win-win proposition.

\section{Conclusions and Future Work}

In this paper we presented ways of measuring and reducing the extractiveness of abstractive generative summaries while increasing their overall quality.

We showed that while one of the best generative summary architectures - the PGNet - alleviates other ills of summaries, including repetition and out of vocabulary words, it suffers from a highly extractive nature.
We outlined why previous methods found in plagiarism detection are not well-suited for evaluating generative summaries and proposed an alternative extractiveness measure. We then showed that this extractiveness measure is correlated with human judgment, with the ground truth being close to the minimum extractiveness values.

We then presented an alternative decoding mechanism which can be applied to any abstractive framework which uses beam search. The method leverages multiple recent developments, complementing the diverse beam search with a diverse summary combination mechanism. The latter is based on similarity measures computed using sentence embeddings.

We showed the advantages of this diverse summary generation method. It not only reduce the extractiveness of the architecture, measured in two separate ways, but that it also improves the overall ROUGE score by a significant 2\%.

Finally, we believe that this method opens the door to several directions of research for reducing the extractiveness of abstractive framework alongside improving their overall performance. 
One one hand, selecting the candidates is a crucial step as doing it optimally can further improve the ROUGE score. 
One the other hand, this method could be experimented on other abstractive summarization frameworks, in order to reduce their extractiveness while improving their overall ROUGE score.

\appendix
\bibliographystyle{named}
\newpage
\bibliography{ijcai18}

\end{document}